\documentclass[runningheads]{llncs}
\usepackage[T1]{fontenc}
\usepackage{caption}
\usepackage{diagbox}
\usepackage{multirow}
\usepackage{amsmath}
\usepackage{mathtools}
\usepackage{amssymb}
\usepackage{wrapfig}
\usepackage{pgfplots}
\usepackage{pgfplotstable}
\usepackage{epstopdf}
\usepackage{graphicx}
\begin{document}
\title{Enhancing AI Face Realism: Cost-Efficient Quality Improvement in Distilled Diffusion Models with a Fully Synthetic Dataset}
\titlerunning{Enhancing AI Face Realism}
\author{
Jakub Wąsala\orcidID{0009-0003-4808-6608} \and
Bartłomiej Wrzalski\orcidID{0009-0007-0130-3379} \and  \\
Kornelia Noculak\orcidID{0009-0005-8616-4528} \and
Yuliia Tarasenko\orcidID{0009-0004-3027-6742} \and \\
Oliwer Krupa\orcidID{0009-0009-7546-6140} \and 
Jan Kocoń\orcidID{0000-0002-7665-6896} \and \\
Grzegorz Chodak\orcidID{0000-0002-9604-482X}}
\authorrunning{J. Wąsala et al.}
\institute{Department of Artificial Intelligence, \\
Wroclaw University of Science and Technology, \\ 
Wroclaw, Poland \\
\email{\{254554, 254566, 254580, 283338, 254027\}@student.pwr.edu.pl}, \\
\email{jan.kocon@pwr.edu.pl}, \email{grzegorz.chodak@pwr.edu.pl}}
\maketitle              % typeset the header of the contribution

\begin{abstract}
This study presents a novel approach to enhance the cost-to-quality ratio of image generation with diffusion models. Our solution introduces a fully synthetic pairwise dataset of images from distilled and baseline versions of a model (i.e. FLUX.1-\textit{schnell} and FLUX.1-\textit{dev}). We hypothesize that differences between distilled and baseline models are consistent and, therefore, learnable within a specialized domain, like portrait generation. Then, we train an image-to-image translation head.

The proposed method works without requiring manual annotations or real reference photos. Using two sets of low- and high-quality synthetic images, our model is trained to refine the output of a distilled generator (e.g., FLUX.1-$schnell$) to a level comparable to a baseline model like FLUX.1-$dev$, which is more computationally intensive.

We train and compare multiple variations of our image-to-image model, both pairwise (U-Net) and non-pairwise (CycleGAN, ESA-CycleGAN).

Our results show that the pipeline, which combines a distilled version of a large generative model with our enhancement layer, delivers similar photorealistic portraits to the baseline version with up to an 82\% decrease in computational cost compared to FLUX.1-$dev$. This study demonstrates the potential for improving the efficiency of AI solutions involving large-scale image generation.

\end{abstract}

\newpage
\begin{figure}
\begin{center}
\includegraphics[width=\textwidth]{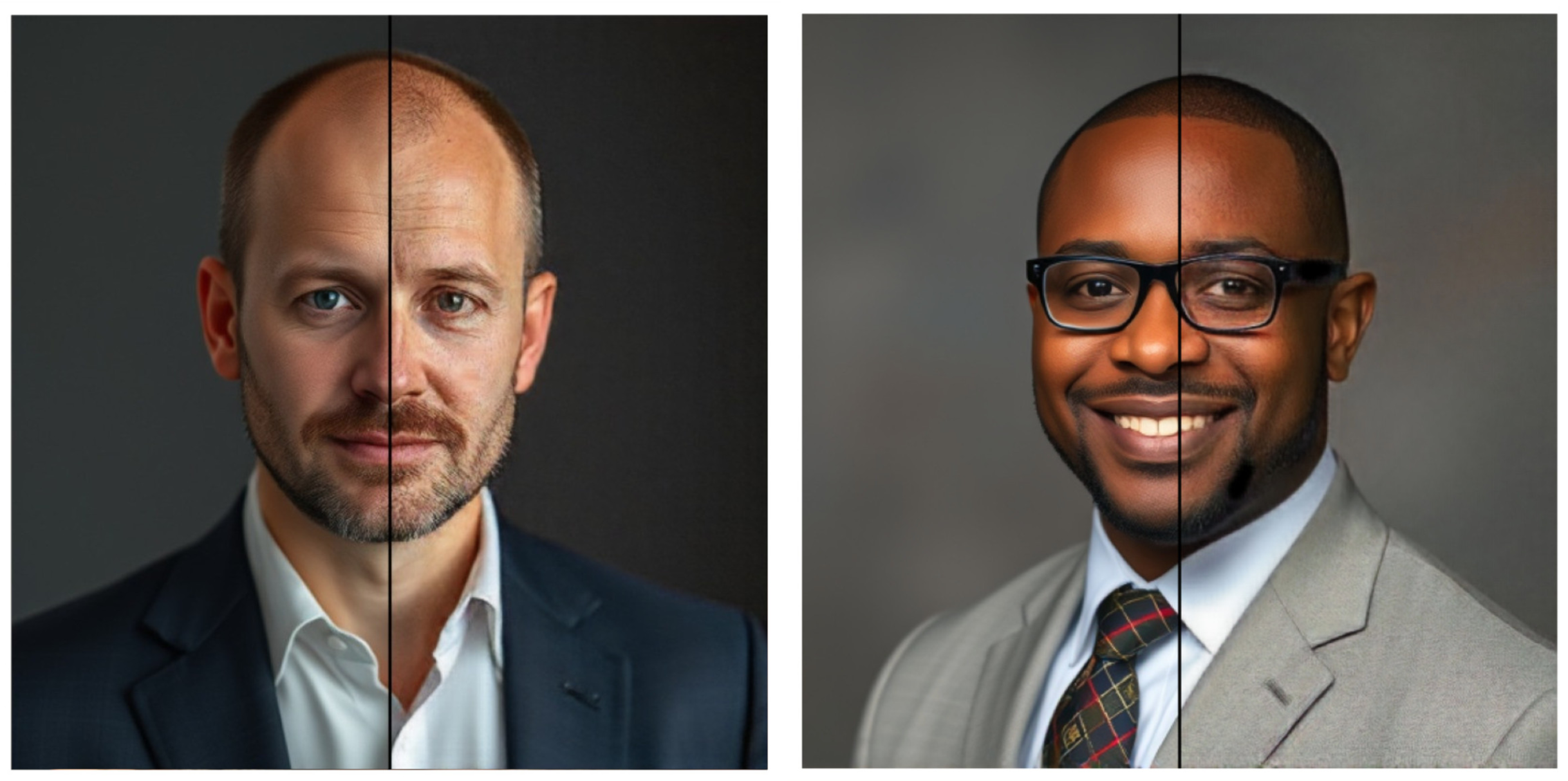}
\caption{\textbf{Left sides:} example input images from FLUX.1-\textit{schnell}. \textbf{Right sides:} outputs from our model based on ESA-CycleGAN. Our model adds skin details, enhances hair, and improves the reflection of eye pupils, contributing to the overall realistic feel of a portrait.} \label{fig1}
\end{center}
\end{figure}
\vspace{-20pt}
\section{Introduction}
Deep generative models estimate data distributions to synthesize realistic images. However, as statistical models, they are limited by a finite set of parameters, making it difficult to capture real-world subjects' full complexity and irregularities. This limitation is particularly evident in portrait imagery generation, where even state-of-the-art models sometimes struggle to reproduce fine details with perfect accuracy \cite{borji2023qualitative}.

In recent years, models such as StyleGAN \cite{Karras}, BigGAN \cite{Brock}, DALL-E \cite{Dalle}, and FLUX.1 \cite{blackforestlabs} have made remarkable progress in generating high-fidelity synthetic images. These advancements have facilitated applications in entertainment, advertising, and professional avatar creation \cite{nautiyal2023intersection}. However, their deployment remains constrained by high computational costs, requiring extensive resources for both training and inference \cite{Dhariwal}. This poses challenges for practical implementation, especially in resource-limited environments.

To this end, lightweight generative models such as FLUX.1-\textit{schnell} \cite{blackforestlabs} have been introduced to reduce computational demands while preserving image quality. However, these models often struggle more with visible imperfections --- including unnatural lighting, texture inconsistencies, and unrealistic facial details --- limiting their effectiveness in professional applications.

Our proposal bridges the gap between resource-efficient distilled models and the quality of their higher-quality counterparts by introducing a dedicated image-to-image (I2I) model trained to recover lost details. As a motivation example, we showcase photorealistic portrait generation using the recent diffusion model FLUX.1. We observe that images generated by its advanced revision, FLUX.1-\textit{dev}, exhibit finer details in hair, skin, and eyes compared to those produced by FLUX.1-\textit{schnell}, a timestep-distilled variant. On the other hand, our experiments show that \textit{schnell} is 82\% more computationally efficient, requiring fewer denoising steps to converge.

To validate our approach, we propose and then test two key hypotheses: \textbf{(1)} the imperfections of a distilled model are consistent within a specialized domain (e.g., portrait photography) and thus can be learned by an I2I model, \textbf{(2)} the combined pipeline --- comprising the distilled model and our I2I head --- is still significantly faster than the full-scale model.

\begin{wrapfigure}{r}{0.5\textwidth} % 'r' for right placement
    % Adjust negative space to move the figure up
    \vspace{-20pt}
    \centering
    \includegraphics[width=0.45\textwidth]{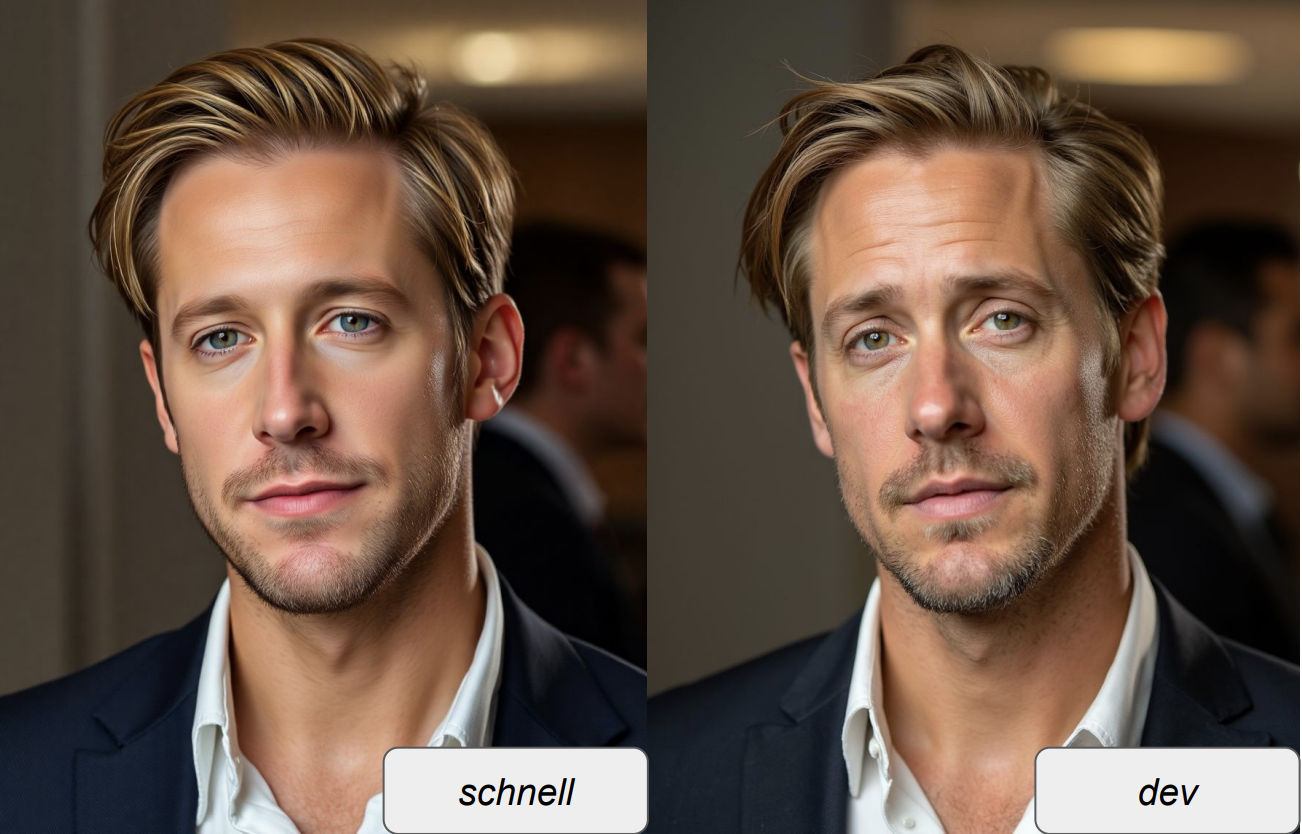} % Keep width slightly smaller than 0.5\textwidth for spacing
    \caption{Example pair of FLUX.1-\textit{schnell} and FLUX.1-\textit{dev} images from our dataset. We observe that the \textit{dev} variant generates portraits featuring more details of skin, hair, and eyes, leading to a more photorealistic look.}
    \label{fig-dataset}
    \vspace{-10pt}
\end{wrapfigure}
A crucial factor in training an effective I2I model is the dataset's quality. We introduce a novel approach to fully synthetic dataset generation to address this. Our method operates pairwise, producing image pairs featuring the same subject and composition but using different FLUX.1 revisions. This ensures controlled variations in quality and photorealistic detail, forming a strong training set for learning quality restoration. We also demonstrate that a simple prompt engineering technique can significantly enhance dataset diversity across race, ethnicity, gender, and age.

We also compare our work to existing solutions aiming at improving the quality of images generated by diffusion models. Typical ways to influence the outputs of a model include prompt engineering and parameter-efficient fine-tuning (PEFT) methods, such as LoRA \cite{hu2022lora}. We note that in comparison with those methods, our approach is technically model-agnostic, as one can forward any image through our model, regardless of its source.

Our contributions are as follows. \textbf{(1)} We demonstrate that, given two versions of a model --- baseline and distilled --- and a specialized domain, such as portrait generation, it is possible to train an image-to-image translation model that effectively mitigates some of the quality degradation introduced by the distilled model. \textbf{(2)} We show that using a distilled version of a generative model with our I2I head leads to 82\% save of computation time. \textbf{(3)} We propose a novel approach to dataset generation involving a tailored prompt engineering technique that results in a fully synthetic, pairwise, and diverse dataset. \textbf{(4)} We use our approach to generate a dataset of synthetic portraits of 280,000 images. \textbf{(5)} We compare different approaches to I2I training, including supervised paired methods based on UNet, and unpaired ones based on CycleGAN \cite{Zhu2017UnpairedIT,Wang2021ESACycleGANEF}.

% - Autorska metoda generowania zbiorów danych
% - Autorska metoda urealniania zdjęć, szybsza niz metody referencyjne
% - Autorska metoda oceny efektywności systemu?
% - Srodowisko anotacyjne jako nasza kontrybucja
% - Autorski zbiór danych
% - Framework
% - Model-agnostic?
% - Uzyteczne produkcyjnie w generowaniu portretow?
% - Dodanie metryk oceny ludzkiej

\section{Related Work}
The rapid development of generative models has significantly improved the quality and realism of synthesized images. While diffusion models have set new standards in photorealistic image generation due to their iterative refinement process \cite{ho2020denoising,Dhariwal}, their high computational cost limits their practical deployment \cite{song2021,guo2022}. This has led to an increasing focus on efficient alternatives, including model distillation \cite{hinton2015}, quantization \cite{jacob2017}, and hybrid fine-tuning approaches \cite{houlsby2019}. Despite these optimizations, distilled and quantized models often introduce artifacts, such as unnatural lighting and texture inconsistencies \cite{rombach2022high,xia2023diffi2iefficientdiffusionmodel}. Addressing these issues is critical for applications requiring high-fidelity image synthesis.

\subsection{Diffusion-Based Image-to-Image Translation}
Diffusion models, such as DDPM \cite{ho2020denoising} and DDIM \cite{song2021}, have revolutionized image synthesis by leveraging iterative denoising processes to generate high-quality images. Compared to GANs, they offer superior realism and diversity but at the cost of high computational demands.

One approach to improving efficiency in I2I tasks is the DiffI2I model, which incorporates a compact prior extraction network and a dynamic transformer to produce accurate translations with reduced computational overhead \cite{xia2023diffi2iefficientdiffusionmodel}. Another challenge is content preservation, which researchers have tackled by disentangling style and content representations, ensuring that generated images maintain the original content while adopting the desired style \cite{kwon2023diffusionbasedimagetranslationusing}. This is particularly relevant for tasks requiring high fidelity to the source image, such as portrait generation.

Recent research has also explored alternative I2I models, including Latent Consistency Models (LCMs) \cite{luo2023}, which accelerate diffusion processes while maintaining quality. Additionally, GAN-based approaches, such as pix2pix \cite{isola2018} and CycleGAN \cite{ZhuPark}, have demonstrated effectiveness in translating images across domains, albeit with challenges in preserving fine details. ControlNet \cite{zhang2023} has further enhanced diffusion-based I2I tasks by providing more precise control over image modifications.

\subsection{FLUX: A Computationally Efficient Generative Model}
FLUX is a recently proposed generative model optimized for computational efficiency \cite{blackforestlabs}. Its architecture is designed to scale effectively while maintaining high-quality image synthesis. Unlike traditional diffusion models, FLUX aims to reduce computational overhead by incorporating lightweight neural components that facilitate faster inference. This makes it particularly suitable for applications requiring real-time generation, such as avatar creation and artistic content generation.

Furthermore, FLUX can be integrated with quantization and fine-tuning techniques to enhance its efficiency while preserving image quality. Recent studies have explored the potential of combining FLUX with parameter-efficient fine-tuning strategies to achieve high-fidelity image synthesis at a lower computational cost \cite{hu2022lora}.

\subsection{Fine-Tuning Techniques}

Fine-tuning pre-trained models is a prevalent strategy to adapt generative models to specific tasks or styles. Two primary methods have emerged: full fine-tuning and parameter-efficient fine-tuning techniques like Low-Rank Adaptation (LoRA).

Comparisons between full fine-tuning and LoRA training suggest that full fine-tuning yields superior results, with reduced overfitting and improved generalization \cite{shuttleworth2024}. However, this approach demands significant computational resources, which may not be feasible in all scenarios. In contrast, LoRA fine-tuning offers a more resource-efficient alternative by introducing low-rank matrices to capture essential adaptations, updating only a subset of parameters, thereby reducing memory usage and training time \cite{hu2022lora}. Additional studies have explored hybrid fine-tuning strategies, such as adapter-based tuning \cite{houlsby2019} and layer-wise modifications \cite{krishnanunni2024}, to balance efficiency and model expressiveness.

Furthermore, specialized fine-tuning techniques such as Realism LoRA \cite{hugovntr_flux_schnell_realism} and DreamBooth \cite{ruiz2023} have been developed to enhance realism in generated images. Realism LoRA refines model weights to produce highly detailed and lifelike outputs, making it valuable for portrait generation and professional photography applications. DreamBooth, on the other hand, enables user-specific fine-tuning by incorporating personalized training data, allowing models to generate images with high identity preservation.

\subsection{Gaps and Contributions}
While existing methods have advanced the fields of diffusion-based I2I translation, fine-tuning, and quantization, challenges remain in achieving high-quality image generation with reduced computational costs. Our study addresses these gaps by introducing a novel approach that leverages a fully synthetic paired dataset derived from baseline and distilled versions of a base model. By training an image-to-image translation head in a supervised manner, our method enhances the output quality of a baseline generator to match that of more computationally intensive models. This strategy eliminates the need for manual annotations or real reference photos, offering a cost-effective solution for photorealistic portrait generation.

In summary, our approach contributes to the existing body of work by providing an efficient framework that combines supervised learning and image-to-image translation to improve the quality of diffusion models without incurring significant computational expenses.

\section{Dataset}
The work presents a novel approach to generating a diverse dataset of portraits. Its main idea was to acquire almost identical pairs of images. They were intended to be semantically coherent and to distinguish only by details such as a more natural skin texture and a better-depicted beard or hair --- typical to FLUX.1-\textit{dev}. Thanks to this technique, there was no need for an outsourced dataset, as we used it to facilitate the assembly of 280,000 portraits.

\subsection{Preparation of the dataset}
For the purpose of preparing the dataset, we used FLUX.1-\textit{schnell} and FLUX.1-\textit{dev}. Both of these models are open-weight and available through the HuggingFace platform. The former is a distilled version of the latter, working at an inferior quality. To begin with, we generated 10,000 samples using the prompt ``A professional business portrait'' with a resolution of 512$\times$512 pixels. However, we observed a lack of diversity in the generated images -- for example, there was a noticeable bias toward masculine subjects.

In the following phase, to increase the diversity of the collection, we applied prompt engineering by enriching the prompts with supplemental names and surnames. For preparing the names, we utilized the IMDb Actors and Movies database. By using a subset of this collection, we generated close to 140,000 images with FLUX.1-\textit{schnell} with the prompt ``A professional portrait of [FULL NAME]'', where each name corresponded to the individual actor in the IMDb database. We hypothesize that the model, even if not trained to properly depict specific individuals based on name, may associate names with diverse cultures.

Subsequently, using the FLUX.1-\textit{dev} model, we produced target images in I2I mode with the identical predefined prompts. We obtained satisfactory results during model inference with a 3.0 guidance scale, 0.7 prompt strength, and 50 inference steps. Beyond higher quality, the resulting samples presented improved realism. Using this approach, a dataset of paired images was acquired, consisting of 280,000 images. Pipeline of the formation of the data is presented in Figure \ref{unet_pipeline}.

\begin{figure}
    \centering
    \includegraphics[width=1\textwidth]{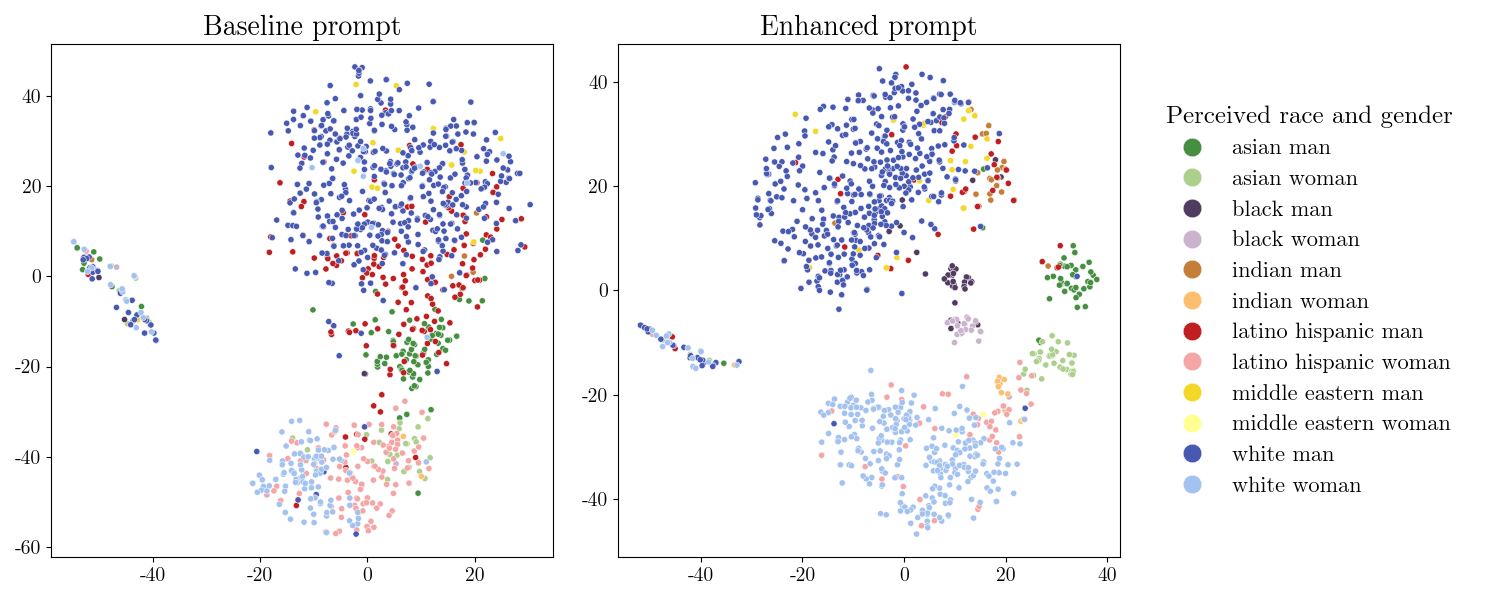}
    \caption{t-SNE visualization of face embeddings. Representations were acquired using the ArcFace model \cite{deng2019arcface}.}
    \label{fig:tsne}
\end{figure}

\begin{table}
    \centering
    \caption{Distribution of attributes --- perceived gender, age, and dominant ethnicity --- in sets based following the baseline prompt and enriched prompt.}
    \begin{tabular}{|l|c|c|c|}
        \hline
        \textbf{Prompt} &  \shortstack{\textbf{Perceived Gender} \\ M : F}  & \shortstack{\textbf{P. Age} \\ 18-30 : 31-50 : 50+} & \shortstack{\textbf{P. Dominant Ethnicity} \\ Asian : Black : Indian \\ Latino : Middle Eastern : White} \\
        \hline
        Baseline & 0.73 : 0.27 & 0.61 : 0.39 : 0.0 & 0.12 : 0.01 : 0.01 : 0.26 : 0.02 : 0.59 \\
        Enhanced & 0.60 : 0.40 & 0.40 : 0.58 : 0.02 & 0.10 : 0.05 : 0.04 : 0.08 : 0.02 : 0.7 \\
        \hline
    \end{tabular}
    \label{tab:diversity-dataset}
\end{table}

\subsection{Analysis of Dataset Diversity}\label{sec:dataset_diversity}
As mentioned previously, we have increased the diversity of portraits by using prompt engineering. In pursuit of this, we used a T2I FLUX.1-\textit{schnell} model. The baseline was a set of images using the prompt ``A professional business portrait'' with a random seed. It was compared with a set that included first and last names in the prompt.

The analysis considered facial attributes such as gender, age, and ethnicity. To perform it, we used a framework provided by the DeepFace library \cite{serengil2021lightface}. Results shown in Table \ref{tab:diversity-dataset} demonstrate that employment of prompt engineering technique led to increased equality in perceived gender occurrences and increased participation of previously underrepresented racial groups, especially Black and Indian. Random samples are presented using t-SNE in Figure \ref{fig:tsne}.

\section{Architecture}
\subsection{Supervised Pairwise Approach \label{sect:arch-unet}}
Our initial approach to model design was based on a \textbf{U-Net} architecture. The model was intended to operate in an I2I fashion, taking an image generated by a lower-quality model as input and transforming it into an image with enhanced quality, returned by a more advanced model.

The base U-Net architecture was extended with residual connections, which improve gradient flow and training stability, and CBAM (Convolutional Block Attention Module) blocks, which allow the model to focus more effectively on important image features. The network had a depth of six layers, enabling it to capture local and global dependencies within the image.

\begin{figure}
\centerline{\includegraphics[width=0.8\linewidth]{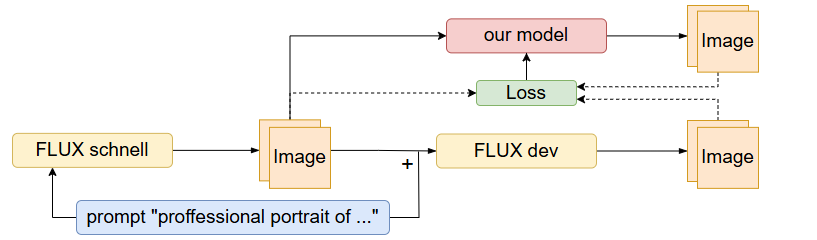}}
\caption{Pipeline of our solution, including the integration of FLUX.1-\textit{schnell} and our image-to-image model.}
\label{unet_pipeline}
\end{figure}

\begin{figure}
\centerline{\includegraphics[width=0.7\linewidth]{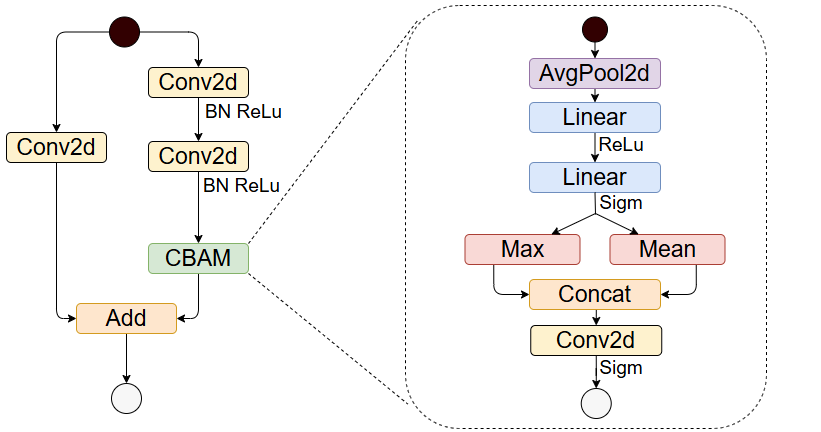}}
\caption{Single block of our image-to-image U-Net model.}
\label{unet_arch}
\end{figure}

The loss function was formulated as the sum of two components: a gradient loss computed between the input and output images and a perceptual loss (LPIPS) measured between the target and generated images. Formally, the function was defined as:

\begin{equation}
    loss = GRAD(x, z) + LPIPS(y, z)
\end{equation}

where $x$ represents the input image, $y$ the target image, and $z$ the image generated by the model. The learning rate was set at $5e^{-5}$, ensuring stable and effective training.

The enhancement model operates as part of a two-step pipeline. First, FLUX.1-\textit{schnell} generates a synthetic portrait. The generated image, which often exhibits imperfections, is then processed by our model to achieve high-quality photorealism. This streamlined approach balances computational efficiency with output quality, making it practical for real-world applications. Figure \ref{unet_pipeline} illustrates the general pipeline of our method, including the integration of FLUX.1-\textit{schnell} and our image-to-image model.

The architecture of our image-to-image model is shown in Figure \ref{unet_arch}. It highlights the detailed design of the U-Net structure, residual connections, and the CBAM attention mechanism.

\subsection{Unsupervised Non-Pairwise Approach \label{sect:arch-cyclegan}}

In addition to the supervised U-Net-based techniques outlined previously, we also explore \emph{unsupervised non-pairwise} methods for image-to-image translation. Unlike supervised approaches, which require paired training samples, these techniques rely on two separate collections of images — one set per domain — without needing explicit one-to-one correspondences. In our case, we define domain \textit{A} as portraits generated by FLUX.1-\textit{schnell}, while domain \textit{B} is comprised of images produced by FLUX.1-\textit{dev}.

A leading example of such non-pairwise translation is \textbf{CycleGAN} \cite{Zhu2017UnpairedIT}, whose core design features two generators: \textbf{(1)} $G$ that converts images from domain $A$ to domain $B$, and \textbf{(2)} $F$ that handles the reverse mapping from $B$ back to $A$. Each generator is trained in tandem with a discriminator: \textbf{(1)} $D_A$ that learns to differentiate real $A$-domain images from generated ones, \textbf{(2)} $D_B$ that discerns real $B$-domain images from synthesized outputs.

A distinguishing element of CycleGAN is the \emph{cycle consistency loss}, ensuring that an image transformed to the opposite domain and then back again remains close to the original. Formally:
\begin{equation}
\mathcal{L}_{\text{cycle}}(G, F) \;=\;
\mathbb{E}_{x\sim A}\bigl[\|F(G(x)) - x\|\bigr]
\;+\;
\mathbb{E}_{y\sim B}\bigl[\|G(F(y)) - y\|\bigr].
\end{equation}
This term prevents the networks from altering content arbitrarily to achieve a plausible look in the target domain.

CycleGAN combines the above term with adversarial objectives for both generators:
\begin{equation}
\mathcal{L}_{\text{total}}
=
\mathcal{L}_{\text{GAN}}(G,D_B)
+ \mathcal{L}_{\text{GAN}}(F,D_A)
+ \lambda_{\text{cycle}} \cdot \mathcal{L}_{\text{cycle}},
\end{equation}
where $\lambda_{\text{cycle}}$ is a key hyperparameter balancing fidelity to the original image against realism in the transformed domain.

To further refine the translation quality, we consider \textbf{ESA-CycleGAN} \cite{Wang2021ESACycleGANEF}, which integrates an Enhanced Spatial Attention (ESA) module into each generator. This module amplifies important regions of the image through: \textbf{(1)} channel reduction and spatial pooling, \textbf{(2)} convolutional analysis of the downsampled features, \textbf{(3)} bilinear upsampling to restore original dimensions, \textbf{(4)} element-wise multiplication to spotlight key structures.

These attention-based enhancements help preserve fine details, particularly in complex or texture-rich images. Empirical evaluations suggest that ESA-CycleGAN improves structural similarity (SSIM) scores and mitigates common artifacts associated with plain CycleGAN. The following sections provide a detailed comparison between the two models, highlighting where the attention-driven approach yields the greatest gains.

\section{Experiments}
\subsection{Supervised Pairwise Training}
We trained a U-Net-like architecture described in \ref{sect:arch-unet}. The training was performed on a fraction (1\%) of the dataset with 512$\times$512 input size. This approach allowed us to quickly run multiple tests with different losses or architectural modifications to compare their achieved improvements.

All training was performed on a single NVIDIA A100 GPU with early stopping to 10 epochs on validation loss. Single training duration was between 20 and 60 minutes, depending on the parameters.

We sample probes and perform human evaluations to evaluate the model's effectiveness and measure whether we improve. We found out that using standard metrics for image quality, like Structural Similarity Index (SSIM) and Peak Signal-to-Noise Ratio (PSNR), does not return reliable values and returns the same values for blurry and sharp images.

To avoid blurry images, we used a combined loss function that measures the structural difference with the input image and the perceptual difference with the target image. We employed Gradient Loss to capture the structural difference between the input and output images, as without it, the generated images would appear unsharp. This issue arises because the target images are not perfectly aligned or mapped with the input images.

Although we incorporated Gradient Loss to improve sharpness, we observed that L1 loss produced similar results in practice.

Our experiments fell short due to artifacts produced by the U-Net-like network. While we achieved sharp and enhanced images, we observed net-like patterns in hairlines and beards, particularly noticeable on light or grey hair. These artifacts negatively impacted the overall realism of the generated images.

\subsection{Unsupervised Non-pairwise Training}
In this experiment, we trained and tested CycleGAN-based approaches introduced in \ref{sect:arch-cyclegan}. This included a classic CycleGAN and its modification, ESA-CycleGAN.

Initial trainings were run on a small fraction (0.5\% to 2\%) of the dataset with 256$\times$256 input image size. This allowed us to quickly examine different values of hyperparameters, such as lambda cycle, learning rate, or batch size. The second training phase involved a smaller set of hyperparameters combinations, a 20\% subset of the entire dataset, and 512$\times$512 input size. We did not rerun training on 100\% of the data since we were already satisfied with the outputs from this phase.

All training experiments were conducted on an HPC cluster with 16 NVIDIA A100 GPUs distributed evenly across four nodes. A final run in this setup took 12 hours to complete 190 epochs.

We compare each training's best CycleGAN loss to select the best setup. We also note SSIM between the input image (SSIM's reference) and CycleGAN's full cycle output (SSIM's target).

\begin{center}

\begin{minipage}{0.45\linewidth}
    \centering
    \captionsetup{type=table}
    \captionof{table}{CycleGAN results.}
    \label{tab:cyclegan_results}
    \begin{tabular}{|r|c|c|c|c|c|c|}
    \hline
    \diagbox{$\lambda_{cyc}$}{\textbf{LR}} & \multicolumn{2}{c|}{$10^{-4}$} & \multicolumn{2}{c|}{$2 \times 10^{-4}$} & \multicolumn{2}{c|}{$3 \times 10^{-4}$} \\
    \cline{2-7}
             & $\mathcal{L}$ & \textbf{SSIM} & $\mathcal{L}$ & \textbf{SSIM} & $\mathcal{L}$ & \textbf{SSIM} \\
    \hline
     10  & 0.96 & \textbf{0.96} & 0.98 & 0.95 & 1.59 & 0.68 \\  
     5   & 0.85 & 0.95 & 1.02 & 0.95 & 1.44 & 0.75 \\  
     2   & \textbf{0.72} & 0.95 & 0.90 & \textbf{0.96} & 1.36 & 0.80 \\  
    \hline
    \end{tabular}
\end{minipage}
\hfill
\begin{minipage}{0.45\linewidth}
    \centering
    \captionsetup{type=table}
    \captionof{table}{ESA-CycleGAN results.}
    \label{tab:esa_cyclegan_results}
    \begin{tabular}{|r|c|c|}
    \hline
    \diagbox{$\lambda_{cyc}$}{\textbf{LR}} & \multicolumn{2}{c|}{$10^{-4}$} \\
    \cline{2-3}
             & $\mathcal{L}$ & \textbf{SSIM} \\
    \hline
     10  & 1.22 & 0.92  \\  
     5   & 0.77 & \textbf{0.96}  \\  
     2   & \textbf{0.61} & 0.95  \\  
    \hline
    \end{tabular}
\end{minipage}

\end{center}

Tables \ref{tab:cyclegan_results} and \ref{tab:esa_cyclegan_results} present the results from the final training runs. The lowest GAN loss was observed for the ESA-CycleGAN variant with $\lambda_{cyc} = 2$. SSIM proved useful for preliminary sanity checks, as values significantly below 0.9 indicated poor-quality results. However, it rapidly converged to approximately 0.95 in higher-quality runs, limiting its effectiveness for more fine-grained performance comparisons.

\subsection{Computational Complexity Assessment}
In this experiment, we evaluate the hypothesis that integrating our model preserves the efficiency benefits of a computationally cheaper distilled backbone. To this end, we run base FLUX.1-\textit{dev} and FLUX.1-\textit{schnell}. Then, FLUX.1-\textit{schnell} is run with an additional I2I head --- in this case, its ESA-CycleGAN variant --- with the same configuration. We repeat the procedure for multiple image sizes.

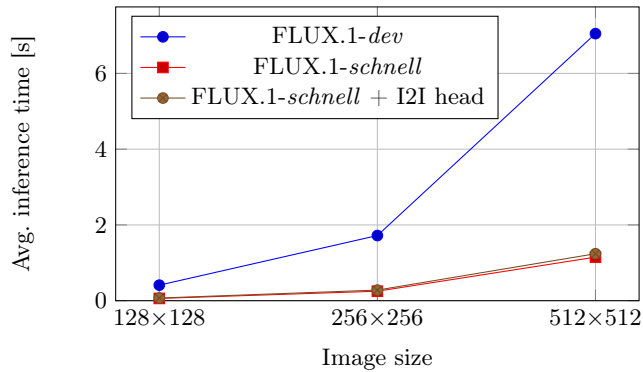
\begin{figure}
    \centering
    \begin{tikzpicture}
        \begin{axis}[
            xlabel={Image size},
            ylabel={Avg. inference time [s]},
            symbolic x coords={128$\times$128,256$\times$256,512$\times$512}, % Treat x-values as categorical
            xtick=data, % Use the symbolic x values for ticks
            xticklabel style={align=center}, % Center align tick labels
            legend pos=north west,
            grid=major,
            ymin=0,
            width=0.7\textwidth,
            height=0.45\textwidth
        ]
        
        % FLUX-dev
        \addplot coordinates {(128$\times$128,0.41) (256$\times$256,1.72) (512$\times$512,7.05)};
        \addlegendentry{FLUX.1-\textit{dev}}

        % FLUX-schnell
        \addplot coordinates {(128$\times$128,0.06) (256$\times$256,0.25) (512$\times$512,1.15)};
        \addlegendentry{FLUX.1-\textit{schnell}}

        % FLUX-schnell + Ours
        \addplot coordinates {(128$\times$128,0.07) (256$\times$256,0.28) (512$\times$512,1.24)};
        \addlegendentry{FLUX.1-\textit{schnell} + I2I head}

        \end{axis}
    \end{tikzpicture}
    \caption{Comparison of average inference time [s] for different pipelines.}
    \label{fig:time_plot}
\end{figure}

Figure \ref{fig:time_plot} highlights the efficiency of our I2I head, demonstrating a significantly lower computational overhead compared to diffusion-based backbones, FLUX.1-\textit{dev} and FLUX.1-\textit{schnell}. At a resolution of 512$\times$512, our I2I head achieves a processing speed of 11.2 images per second, whereas FLUX.1-\textit{schnell} generates only 0.25 images per second. This results in our proposed pipeline achieving an average speedup of 82\% over the standalone FLUX.1-\textit{dev}.

\subsection{Quality Assessment}
To assess how our enhancement model brings the images closer to what the reference model would output, we calculate Fréchet inception distance (FID) \cite{heusel2017gans} twice --- once with FLUX.1-\textit{schnell} as reference, and once with FLUX.1-\textit{dev} as reference. We aim at maximizing the distance from the source dataset to \textit{schnell} images while minimizing the distance to reference \textit{dev} images. To this end, we introduce a FID$_{diff}$ metric, which is simply a subtraction of FID$_{dev}$ from FID$_{schnell}$. We compare different variations of enhanced FLUX.1-\textit{schnell}: \textbf{(1)} pairwise trained I2I head (ours), \textbf{(2)} non-pairwise trained I2I head (ours), \textbf{(3)} \texttt{flux-lora-realism} \cite{hugovntr_flux_schnell_realism} adaptation available at the Huggingface platform. Results are presented in the Table \ref{tab:metrics}.

We also employed an additional non-reference metric, namely CLIP-IQA \cite{wang2023exploring}. It works by evaluating the overall quality of the images without comparing them directly to a reference dataset.

\begin{table}
\centering
\caption{Quality comparison of different approaches to I2I enhancement.}\label{tab:metrics}
\begin{tabular}{|l|c|c|c|c|}
\hline
\textbf{Model} & \textbf{FID}$_{schnell} $ & \textbf{FID}$_{dev}$ & \textbf{FID}$_{diff}$ & \textbf{CLIP-IQA} \\
\hline
FLUX.1-\textit{schnell} & - & 0.37 & - & 0.35 \\
FLUX.1-\textit{dev} & 0.37 & - & - & 0.36 \\
\hline
LoRA Realism & 0.32 & 0.59 & -0.27 & 0.34\\
Ours (pairwise) & 0.54 & 0.53 & 0.01 & 0.34\\
Ours (non-pairwise) & 0.75 & \textbf{0.34} & \textbf{0.41} & 0.35\\
\hline
\end{tabular}
\end{table}

Our non-pairwise method produces images perceptually closer to those generated by base FLUX.1-\textit{dev} than to those from base FLUX.1-\textit{schnell}. In contrast, the pairwise-trained model yields outputs that exhibit similar perceptual distances to both reference sets. Notably, the LoRA-based approach results in a negative FID$_{diff}$, indicating that its outputs are perceptually closer to FLUX.1-\textit{schnell} without adaptation than to FLUX.1-\textit{dev}. Our secondary metric, CLIP-IQA, proved ineffective for quality assessment, as it produced similar scores regardless of the target. More critically, it fails to reflect the superior quality of FLUX.1-\textit{dev}, with score differences falling within a reasonable error margin.

Importantly, the trends captured by FID$_{diff}$ align well with our perceptual evaluation, reinforcing its reliability as a quality measure.

\section{Discussion}
The study establishes a cost-efficient method to improve portrait photos, testing paired and unpaired approaches. We show that a lightweight I2I enhancement head can be used to bring the outputs of a smaller, distilled diffusion model closer to the general look and feel of a larger, full-scale base model --- while not diminishing the computational save. Unlike existing methods, like Realism LoRA, our model works completely agnostic to the generation method and can be applied at any time later.

The results achieved with CycleGAN and ESA-CycleGAN backbone demonstrate that pairing photos was not necessary to improve realism --- indeed, non-pairwise training delivered even better results. The samples obtained with this approach contained fewer artifacts and featured enhanced details with noticeable detail enhancement.

For the U-Net approach, the presence of net-like pattern artifacts was particularly challenging. The effectiveness of this technique was further hindered by imperfections in the dataset. The images produced by the I2I model would contain slightly different features, such as modified backgrounds, jewelry, or clothing, among others.

While our best model works without pairwise training, we recognize the potential value of our novel approach to dataset construction. The presented pipeline and prompt engineering technique allow controllable distribution of perceived race, gender, and age of subjects, paving the path towards diverse, unbiased synthetic datasets. In future work, the study can be continued by extending the exploration of the methodology as well as expanding the dataset by utilizing other generative models as well (e.g., Stable Diffusion 3.5 Large).

\begin{credits}
\subsubsection{\ackname} Work financed by:   
(1) CLARIN ERIC funded by the Polish Minister of Science, no. 2024/WK/01;
(2) The European Regional Development Fund, FENG programme, no. FENG.02.04-IP.040004/24;
(3) Department of Artificial Intelligence, Wroclaw Tech;
(4) Polish HPC infrastructure PLGrid (HPC Center: ACK Cyfronet AGH): computer facilities and support within grant no. PLG/2024/017768.

\subsubsection{\discintname}The authors have no competing interests to declare that are
relevant to the content of this article.
\end{credits}
\bibliographystyle{splncs04}
\bibliography{bibliography}

\section*{Appendix}
We include more examples of our models, pairwise and non-pairwise, in action.
\vspace{-10pt}
\begin{figure}
\begin{center}
\includegraphics[width=\textwidth]{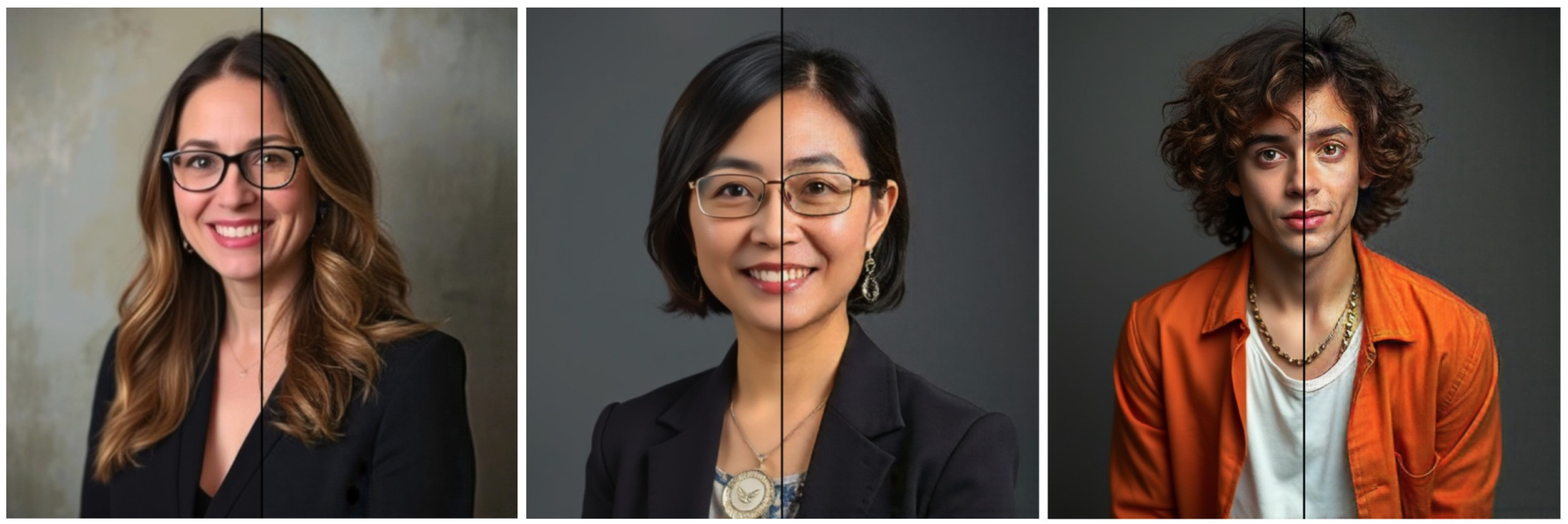}
\includegraphics[width=\textwidth]{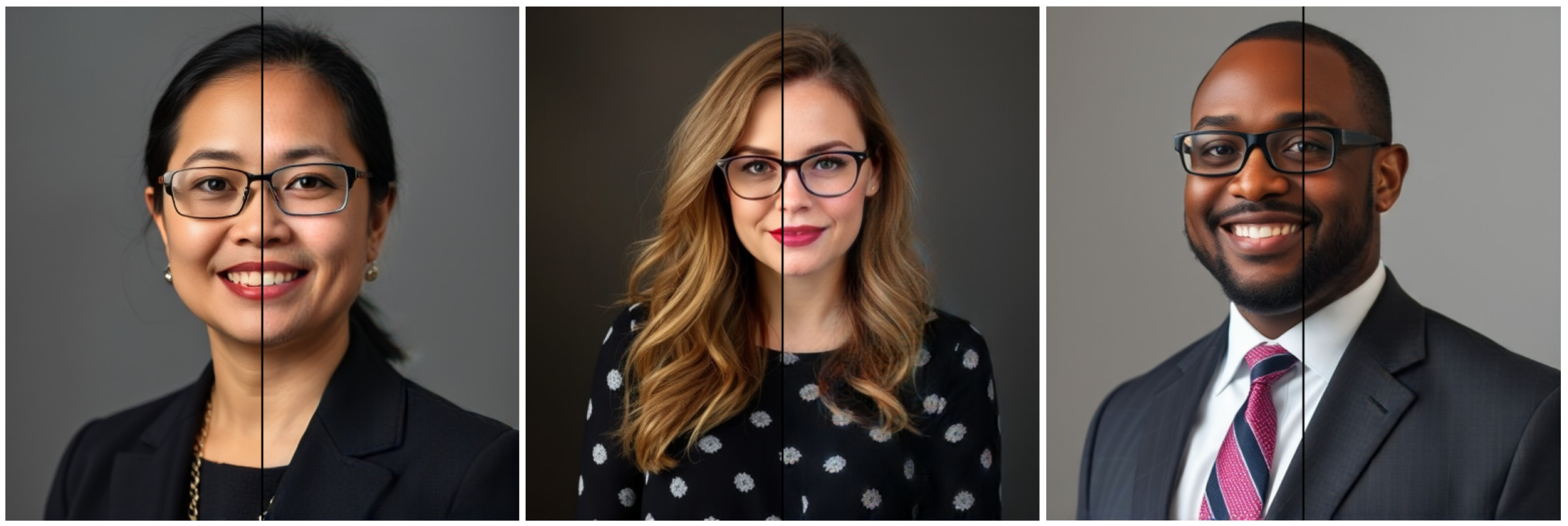}
\caption{\textbf{Left sides:} example input images from FLUX.1-\textit{schnell}. \textbf{Right sides, top row:} outputs from our non-pairwise model based on ESA-CycleGAN. \textbf{Right sides, bottom row:} outputs from our pairwise model based on U-Net.}  \label{fig2}
\end{center}
\end{figure}
\vspace{-10pt}

\end{document}